\documentclass{bmvc2k}

\usepackage{caption} 
\usepackage{algorithm}
\usepackage{algpseudocode}
\newtheorem{thm}{Theorem}
\usepackage{amsfonts}       
\usepackage{nicefrac}       
\usepackage{microtype}      
\usepackage{xcolor}         
\usepackage{mathrsfs}
\usepackage{textcomp}
\usepackage{bbm}
\usepackage{subcaption}
\usepackage{tabularx}
\usepackage{colortbl}

\usepackage{graphicx}
\usepackage{booktabs}
\usepackage{wrapfig}

\usepackage[accsupp]{axessibility}  

\title{Text-Guided Mixup Towards Long-Tailed Image Categorization}

\addauthor{Richard Franklin}{rsamfranklin@gmail.com}{1}
\addauthor{Jiawei Yao}{jwyao@uw.edu}{1}
\addauthor{Deyang Zhong}{dyzhong@uw.edu}{1}
\addauthor{Qi Qian}{qi.qian@alibaba-inc.com}{2}
\addauthor{Juhua Hu}{juhuah@uw.edu}{1}
\addinstitution{
 School of Engineering and Technology\\
 University of Washington\\
 Tacoma, WA
}
\addinstitution{
 Alibaba Group\\
 Bellevue, WA
}
\runninghead{Franklin, Yao, Zhong, Qian, Hu}{Text-Guided Mixup}

\begin{document}

\maketitle

\begin{abstract}
In many real-world applications, the frequency distribution of class labels for training data can exhibit a long-tailed distribution, which challenges traditional approaches of training deep neural networks that require heavy amounts of balanced data. Gathering and labeling data to balance out the class label distribution can be both costly and time-consuming. Many existing solutions that enable ensemble learning, re-balancing strategies, or fine-tuning applied to deep neural networks are limited by the inert problem of few class samples across a subset of classes. Recently, vision-language models like CLIP have been observed as effective solutions to zero-shot or few-shot learning by grasping a similarity between vision and language features for image and text pairs. Considering that large pre-trained vision-language models may contain valuable side textual information for minor classes, we propose to leverage text supervision to tackle the challenge of long-tailed learning. Concretely, we propose a novel text-guided mixup technique that takes advantage of the semantic relations between classes recognized by the pre-trained text encoder to help alleviate the long-tailed problem. Our empirical study on benchmark long-tailed tasks demonstrates the effectiveness of our proposal with a theoretical guarantee. Our code is available at \href{https://github.com/rsamf/text-guided-mixup}{https://github.com/rsamf/text-guided-mixup}.
\end{abstract}

\section{Introduction}
\label{sec:intro}
In recent years, deep learning has made state-of-the-art advancements in computer vision tasks such as image categorization, object detection, and semantic segmentation~\cite{yang2022focal, liu2021swin}. Deep learning models are highly dependent on large-scale and balanced training data, but real-world data are typically class-imbalanced \cite{openlongtailrecognition, van2018inaturalist, NEURIPS2019_621461af}. When training data is abundant for a subset of classes (i.e., head classes) but scarce for the other (i.e., tail classes), the distribution of the data is said to be long-tailed~\cite{zhang2023deep}. Taking image categorization as an example, deep neural networks (DNNs) aim to minimize the empirical risk on the training data by incrementally adjusting the learnable parameters. However, given a long-tailed training data, this happens more on the head-class instances that appear more frequently, augmenting the model's performance bias towards head classes but reducing the model's generalization performance on tail classes~\cite{wang2021longtailed, NEURIPS2019_621461af}.

Long-tailed learning proves to be a significantly challenging task as addressed by many previous studies~\cite{wang2021longtailed, zhang2022self, zhou2020bbn, xu2023learning, DBLP:journals/corr/abs-2111-14745}. Intuitively, under-sampling the head classes and over-sampling the tail classes is a reasonable technique. Although class-level re-sampling or re-weighting can help balance out the data distribution and mitigate the model's performance bias on head classes, these techniques can cause the model's overfitting on tail classes and/or degenerate the performance on head classes~\cite{ren2020balanced}. There is evidently more success in module improvement techniques \cite{zhang2023deep, 7780949, 7780469}, especially those that use ensemble learning \cite{zhang2022self, wang2021longtailed, zhou2020bbn}. There are a number of additional techniques \cite{zhang2023deep} that aim to mitigate the long-tailed problem such as class-level re-margining \cite{NEURIPS2019_621461af}, data augmentation \cite{8388338, chou2020remix}, and transfer learning \cite{DBLP:journals/corr/abs-1803-09014}. However, these methods are still limited by the scarce information found among tail classes. 

\begin{wrapfigure}{r}{0.38\textwidth}
  \begin{center}
    \includegraphics[width=0.33\textwidth]{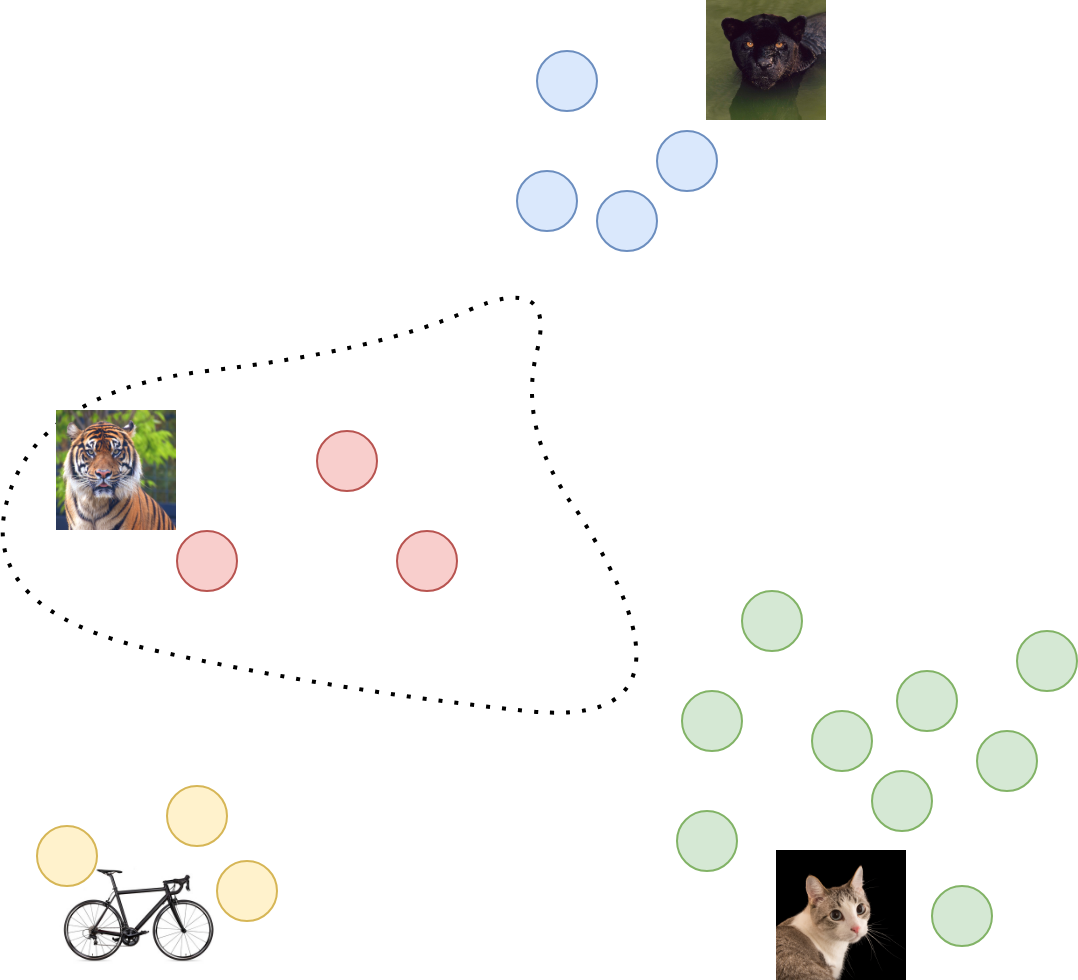}
  \end{center}
  \caption{The decision boundary of `tiger' stretches towards that of `leopard' and `cat' and away from `bicycle' as text-guided mixup allows semantically similar classes to be mixed more frequently.}\label{lfm_effect_boundaries}
\end{wrapfigure}

Recently, vision-language models such as CLIP~\cite{radford2021learning} and ALIGN~\cite{jia2021scaling} have demonstrated good performance in zero-shot classification and few-shot learning \cite{ijcai2021p597}. These models are trained on large-scale data containing image-text pairs that elicit the forming of connections between text and image embedding. By capturing the contrastive locality of image and text features, vision-language models can generalize to unseen categories well, which is a potential information source of tail classes in long-tailed learning. However, existing multi-modal works~\cite{long2022retrieval,tian2022vl,dong2023lpt,jia2022visual} are limited by the general domain knowledge of the CLIP's pre-trained text encoder and must continue linguistic training on the downstream task.

In this work, we propose to leverage the frozen CLIP text encoder to obtain prompt embedding as additional supervision for long-tailed learning in vision tasks.
Considering the observation that semantic relationships between class names (e.g., `tiger' and `cat') correlate with their localities of visual features in vision-language models, we can utilize semantically similar classes to assist the generalization among tail classes (e.g., the head class `cat' can help assist the tail class of `tiger' as shown in Fig.~\ref{lfm_effect_boundaries}). However, the intra-class variance of the tail class can still be ignored. Therefore, we further propose a novel text-guided mixup strategy, named local feature mixup (LFM), to shift the label towards tail classes, so as to alleviate the long-tailed problem. The main contributions of this work are summarized as follows.
\begin{itemize}
    \item We leverage the frozen CLIP text encoder to enhance the performance of long-tailed visual recognition tasks.
    \item We construct a novel mixup technique that takes advantage of the text encoder to boost the performance of tail classes with a theoretical guarantee.
    \item Our extensive experiments on several benchmark long-tailed data demonstrate the effectiveness of our proposal.
\end{itemize}

\section{Related Work}
 
In long-tailed visual recognition, numerous methods have been proposed to boost the performance of tail classes \cite{zhang2023deep}. Module improvement methods including \textbf{ensembling} have shown recent success \cite{zhang2022self, wang2021longtailed, zhou2020bbn, jin2023longtailed}. In mixture of experts, TADE \cite{zhang2022self} and SHIKE \cite{jin2023longtailed} output an aggregation of multiple expert modules, where each expert in TADE strives to perform well in a different training distribution, and each expert in SHIKE focuses on modeling a different depth of image features. Although ensembling can boost performance, these methods are still limited by the scarce information found among instances of the tail classes.

Moreover, \textbf{class re-balancing} such as class-level re-sampling \cite{9694621}, re-weighting \cite{cui2019class} (e.g. Balanced Cross Entropy \cite{ren2020balanced}), and re-margining (e.g., LDAM \cite{NEURIPS2019_621461af}) can adjust the model's attention to classes with a lower sample rate. However, class-balanced sampling or re-weighting can lead to overfitting of the tail classes, under-represent the intra-class variance of the head classes \cite{ren2020balanced, zhang2023deep}, and thus decrease the model's overall performance \cite{sinha2020class}. Alternatively, it can be effective to train a model with meta sampling \cite{ren2020balanced}, in which the optimal sample rate per class is estimated by applying a learnable parameter for each class label. Using this method can slightly avoid the overfitting of tail classes, but finding the optimal parameter or trade-off between class labels for multi-class classification is difficult.

Another instance of success is found through \textbf{pre-training} vision transformers \cite{DBLP:journals/corr/abs-2010-11929, liu2021swin} in an autoencoder setup \cite{kingma2019introduction, higgins2017betavae, samuel2021generalized}. Once the encoder is sufficiently trained, it feeds into a classification layer that is trained using a balanced binary cross-entropy loss \cite{NEURIPS2019_621461af}. However, these methods still lack sufficient performance on the set of tail classes as it is an inert challenge to train deep neural networks for classes with small sample rates. Recently, pre-trained vision-language models like \textbf{Contrastive Language-Vision Pre-training} (CLIP) \cite{radford2021learning} have demonstrated strong zero-shot performance. CLIP embodies multi-modal learning through unsupervised training of image-caption pairs available on the wild web to capture the contrastive locality of image and text features. This makes CLIP more adaptable to new tasks, so that they can be leveraged to make zero-shot predictions, that is, generalize to unseen categories. Thereafter, a pre-trained vision-language model can be further fine-tuned on a downstream task in few-shot learning \cite{alayrac2022flamingo} or long-tailed learning (e.g., RAC \cite{long2022retrieval}, VL-LTR \cite{tian2022vl}, LPT \cite{dong2023lpt}, TeS~\cite{Wang_2023_ICCV}, and VPT \cite{jia2022visual}). However, most of them have been focusing more on the text encoder. For example, VL-LTR~\cite{tian2022vl} requires manually retrieving text descriptions of each class from the Internet to augment the text data in preparation for linguistic training, which is resource expensive, so we instead freeze the text encoder.

\section{The Proposed Method}
\begin{figure*}[ht]
\centering
\includegraphics[width=\textwidth]{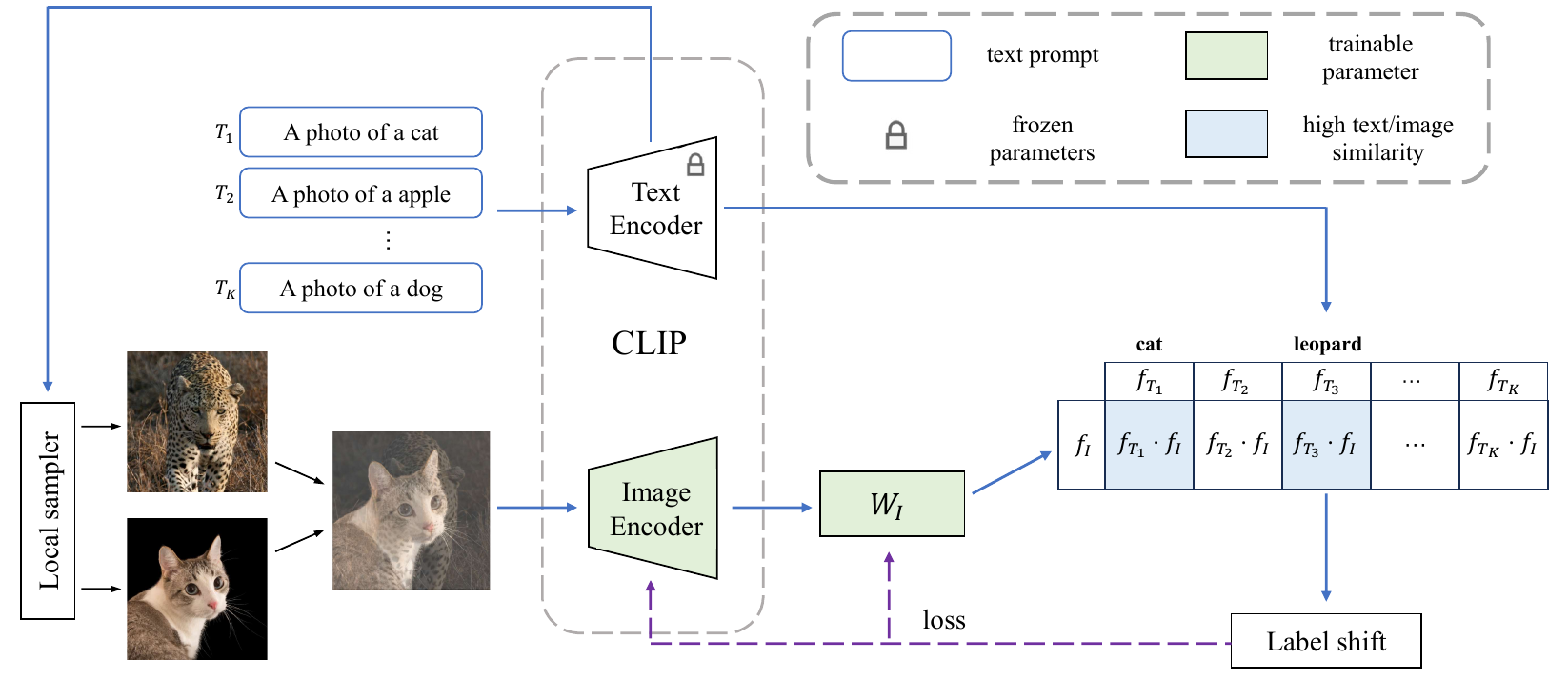}
\caption{The proposed model architecture, in which the text encoder is fixed using the pre-trained model by CLIP~\cite{radford2021learning}. The image encoder will be fine-tuned according the downstream task and $W_I\in \mathbb{R}^{d\times d}$ is appended and learnable.}
\label{model_arch}
 \vspace{-4mm}
\end{figure*}

Given a long-tailed training data $D = \{(x_i,y_i)\}$, $x_i$ is an image associated with its target class $y_i\in \{1, \dots, C\}$. We construct a set of text snippets $T$, where each $T_k$ describes a class label for $k\in \{1, \dots, C\}$. For example, the text snippet describing class name ``dog'' is a tokenized sequence generated from the string as ``a photo of a dog''. 

We feed image and text snippets $T$ to the image and text encoders, respectively, pre-trained by CLIP~\cite{radford2021learning} as shown in Fig.~\ref{model_arch}, for which we denote as $\mathscr{F}_I$ and $\mathscr{F}_T$, respectively. Both of these encoders output feature vectors of size $d$. We denote the output from the text encoder as $f_T$ and $f_T = \mathscr{F}_T (T)$ and allow $f_{T_k}$ to denote the feature vector for class $k$, which does not change during the long-tailed learning. To better separate the tail-class feature embeddings from that of the head-classes following \cite{DBLP:journals/corr/abs-2111-14745}, we append a fully connected layer $W_I \in \mathbb{R}^{d\times d}$ that is learnable to $\mathscr{F}_I$. Thereafter, we can extract the feature vector for each image $x_i$ as $f_I = W_I\mathscr{F}_I(x_i)$. Additionally, we normalize both $f_{T_k}$ and $f_I$ to be of a unit norm.

After obtaining $f_I$ and $f_T$, image classification is performed as shown in Fig.~\ref{model_arch} by computing the cosine similarity between $f_I$ and $f_{T_k}$ for all $k$, and finally, the predicted class label, $\hat{y}$, for each image is computed as $\hat{y} = \operatorname*{arg\,max}_{k \in \{1, \dots, C\}} f_I \cdot f_{T_k}$.

Thereafter, we can adopt a decoupled training approach as suggested by \cite{DBLP:journals/corr/abs-2111-14745} to learn better embeddings for tail-classes compared to joint training. In stage 1, we open $\mathscr{F}_I$ and freeze $W_I$ for training, and in stage 2, we freeze $\mathscr{F}_I$ and open $W_I$. At the beginning, $W_I$ is initialized as the identity matrix with $f_I = \mathscr{F}_I(x)$. However, by minimizing the empirical risk directly based on the training data with a long-tailed distribution, both $\mathscr{F}_I$ and $W_I$ can still be biased to the head classes. Therefore, we propose a novel text-guided mixup technique.

\paragraph{Local Feature Mixup}
A statistical measure of class imbalance in a dataset can be defined as the imbalance factor $\gamma = n_1 / n_C$, where $n_k$ is the number of examples in class $k$ and $n_1\ge n_2\ge ...\ge n_C$ is ordered from high to low, and typically, we have $n_1\gg n_C$. Our main goal is to increase the few-shot accuracy (i.e., those with low $n_k$), while not attenuating the model's accuracy on many-shot classes (i.e., those with high $n_k$). We strive to boost the few-shot accuracy by making two assumptions about the data. First, we assume that classes with low $n_k$ are underrepresented because a few examples may not fully express the complete diversity (or variance) of their associated class. For example, a cat can look different from another cat in terms of their features such as their sizes, their eye colors, and the color/pattern of their furs. When limited to observing a few examples of cats, it is difficult for DNNs to grasp the full range of features that a cat can express. Therefore, we assume that every tail class has a larger intra-class variance than that can be learned from long-tailed data.

Secondly, because both CLIP's image and text encoders map their respective inputs to $d$-dimensional feature vectors, we say that every class can be represented by certain feature space in $\mathbb{R}^{d}$. The pre-trained text encoder already has an understanding of the local relationships between words. For example, words ``frog'' and ``toad'' are close in the language model feature space, since they have similar meanings. Part of our learning objective is to closely align the outputs of our image encoder to the outputs of the pre-trained language model. That is, if we feed an image of a frog and an image of a toad to our image encoder, their extracted feature vectors should be close in proximity as in the text feature space. Therefore, we also assume that if two classes have similar meanings (i.e., nearby in the text encoder's feature space), these two classes also share a subset of visual features and thus should also be nearby within the image encoder's feature space. In the following construction of local feature mixup, we incorporate these two critical ideas separately, that is, local sampling and label shift.

\paragraph{Local Sampling}
Existing mixup strategies often randomly sample $y_i$ and $y_j$ uniformly across the training data \cite{zhang2018mixup, chou2020remix, verma2019manifold}. However, we aim to choose pairs that are semantically related supervised by the pre-trained text encoder. First, we sample an instance from class $y_i$ uniformly across the training data as $p(y = y_i) = \frac{n_i}{\sum^C_{k}n_k}$. Then, we sample another instance from class $y_j$ with probability $p_{ls}(y = y_j | y_i)$ given by Eqn.~\eqref{p_j_i}. 

\begin{equation}
\label{p_j_i}
p_{ls}(y = y_j | y_i) = 
    \begin{cases}
        \frac{ exp( f_{T_i} \cdot f_{T_j} / \tau ) }{ \sum^{C}_{k=1} exp( f_{T_i} \cdot f_{T_k} / \tau ) } & \text{ } i \neq j\\
        0 & \text{o.w.}\
    \end{cases}
\end{equation}
where the hyperparameter $\tau > 0$ controls the temperature scaling on the softmax equation. A lower $\tau$ increases the likelihood that similar class pairs are chosen for mixup, but a too low temperature can lead to oversampling of nearby classes. We set $\tau = 0.05$ for most experiments. Using this strategy, we hope to extend the variance of minor class samples towards neighboring classes as our assumption is that semantically similar classes share a subset of visual features as depicted in Fig.~\ref{lfm_effect_boundaries}.

\paragraph{Label Shift}
Then, we perform mixup by mixing images $x_i$ and $x_j$ sampled through our above local sampling method. With mixing factors $\lambda_x, \lambda_y \in [0,1]$, we propose
\begin{align*}
    \tilde{x}^{LFM} &= \lambda_x x_i + (1-\lambda_x)x_j\\
    \tilde{y}^{LFM} &= \lambda_y y_i + (1-\lambda_y)y_j
\end{align*}
where $y_i, y_j$ are one-hot vectors and factor $\lambda_x$ is chosen randomly from the beta distribution. More importantly, we generate $\lambda_y$ by
\begin{align}
    \label{label_shift}
    \lambda_y = \operatorname{clamp} \left( \lambda_x - \alpha\frac{n_i - n_j}{n_i + n_j}, 0, 1 \right)
\end{align}
where hyperparameter $\alpha \geq 0$ adjusts the intensity of label shift and the resulting value is clamped between $0$ and $1$. In order to expand the margin for tail classes, we shift the decision boundary away from tail classes and towards head classes according to the difference of $n_i$ and $n_j$. For example, if $n_i > n_j$ (i.e., class $y_i$ has more samples than class $y_j$), we shift the target to be more in favor of class $y_j$, thus increasing the model's margin on the class with fewer samples. Algorithms are summarized in the supplementary and we provide a theoretical guarantee for our proposal as follows, while the overall framework is illustrated in Fig.~\ref{model_arch}.

\begin{thm}
Letting $p=n_i/(n_i+n_j)$, $\lambda_y$ can be obtained by balancing the distribution between $x_i$ and $x_j$
\[\lambda_y = \arg\min_{\lambda\in[0,1]} (\lambda - \lambda_x)^2/2+\alpha R(\lambda)\]
where $R(\lambda) = (\lambda-1/2)^2 - (\lambda - p)^2$.
\end{thm}

\paragraph{Remark} The former term constrains that the obtained weight for the label should be close to the weight for the example, while the latter term is a balance regularization to incorporate the prior distribution $p$ between two examples. By minimizing the regularization, it aims to push $\lambda$ from the imbalanced initial distribution to a balanced one. When $p=1/2$, it degenerates to the standard weight for mixup.

\section{Experiments}
To demonstrate the proposed LFM method, following the common practice in long-tailed learning, we use publicly available long-tailed datasets, that is, CIFAR10-LT and CIFAR100-LT \cite{krizhevsky2009learning}, ImageNet-LT \cite{openlongtailrecognition}, and Places-LT \cite{zhou2017places}. 

\paragraph{Experiment Setup}
For CIFAR10/100-LT, we fine-tune CLIP with a single GPU, and for ImageNet-LT and Places-LT, we fine-tune CLIP with three GPUs. Each GPU is an Nvidia GeForce RTX 2080 Ti with 11GB of memory. During training, each GPU receives a batch size of 32, so for ImageNet-LT and Places-LT the effective batch size is 96. Training is performed with a fixed seed to allow for reproduceability. The hyperparameters chosen for LFM are fixed (i.e., $\alpha = 1$, $\tau = 0.05$) for all experiments on CIFAR10-LT, CIFAR100-LT, and ImageNet-LT, while they are adjusted on Places-LT as $\alpha = 1.25$, $\tau = 1.00$ in stage 1 and $\alpha = 1.50$, $\tau = 1.00$ in stage 2, due to the imbalance severity as explained in the next section. Low learning rates were picked to avoid the risk of catastrophic forgetting and losing CLIP's zero-shot performance advantage. 
The detailed hyperparameters used can be found in the supplementary. CLIP's default text prompt template is ``a photo of a \{CLASS\}''. For all experiments, we utilize the default text prompt template provided. 

A model's performance is not necessarily stable across all classes, each with different sample counts, so it is important that we quantify the performance of our model in subdivisions relative to every $n_k$. Across all datasets, we subdivide the resulting model's accuracy into four categories, namely many-shot, medium-shot, few-shot, and overall following \cite{DBLP:journals/corr/abs-1910-09217}. Many-shot classes have $n_k > 100$, medium-shot classes have $20 \leq n_k \leq 100$, and few-shot classes have $n_k < 20$. For each performance category, we report the top-1 accuracy of our model against the balanced validation set for each subdivision of our chosen datasets. 

We compare our proposed method with vision-focused baseline methods and strategies that perform well in tackling the long-tailed problem. We also fine-tune the competitive image encoder (i.e., ViT-B/32) with different existing losses as baselines, i.e., Cross Entropy (CE), Balanced Cross Entropy (BalCE) \cite{ren2020balanced}, Focal \cite{lin2017focal}, Label Distribution Aware Margin (LDAM) \cite{NEURIPS2019_621461af}, and Margin Metric Softmax (MMS) \cite{shu2023clipood}. All losses except CE were proved to be helpful for the class imbalance problem. In summary, we compare with the following baselines based on the pre-trained CLIP~\cite{radford2021learning}: 1) Zero-shot: The pre-trained image and text encoders by CLIP~\cite{radford2021learning} are directly used to do prediction on the balanced test data, in which ViT-B/32 is adopted; 2) CE: Fine-tuned ViT-B/32 using the cross entropy loss; 3) BalCE: Fine-tuned ViT-B/32 using the balanced loss~\cite{ren2020balanced}; 4) Focal: Fine-tuned ViT-B/32 using the Focal loss \cite{lin2017focal}; 5) LDAM: Fine-tuned ViT-B/32 using the loss in LDAM \cite{NEURIPS2019_621461af}; and 6) MMS: Fine-tuned ViT-B/32 with MMS \cite{shu2023clipood}.

\paragraph{CIFAR10/100-LT}
As in the literature, we can create CIFAR10-LT and CIFAR100-LT by taking a subset of the original balanced CIFAR10 and CIFAR100 datasets \cite{krizhevsky2009learning}, and the imbalance factor $\gamma$ is variable. We experiment with multiple imbalance factors in $\{10, 50, 100\}$.

\begin{table}[t]
\centering\small
\caption{CLIP accuracy on CIFAR100-LT with imbalance factor 100, where the best for each is in bold.}
\label{subs_cifar_accuracy}
\begin{tabular}{l|cccc}
\toprule
Methods & Many & Med & Few & All\\
\midrule
Zero-shot
& 63.5 & 60.8 & 61.4 & 62.0 \\
CE
& 79.3 & 67.4 & 53.9 & 67.5 \\
BalCE
& 74.6 & 69.8 & 57.4 & 67.6\\
Focal
& 80.2 & 65.0 & 54.0 & 66.9\\
LDAM
& 81.6 & 70.4 & 58.1 & 70.5\\
MMS
& \textbf{90.3} & 75.2 & 58.1 & 75.2 \\
\hline

LFM + CE
& 81.2 & 79.6 & 68.6 & 77.3\\

LFM + MMS
 & 81.0 & \textbf{81.3} & \textbf{76.5} & \textbf{79.4}\\

\bottomrule
\end{tabular}
\end{table}

First, on CIFAR100 with the imbalance factor of 100, we compare all methods based on CLIP. For our proposal, 
we set ViT-B/32 as the backbone 
and apply LFM with two different losses, i.e., cross entropy and MMS~\cite{shu2023clipood} that is the best in the literature. The comparison results to all baselines are summarized in Table~\ref{subs_cifar_accuracy}. Based on the zero-short performance, we can observe that the pre-trained CLIP can help balance the performance in different categories, which demonstrates the effectiveness of pre-trained vision-language model to alleviate the class imbalance issue. Then, by fine-tuning the pre-trained image encoder, the overall accuracy can be improved. However, due to the severe imbalance, the performance of the tail classes is still lacking even when balanced losses are utilized. Our proposal can help improve the accuracy in all categories, where LFM combined with a loss well-suited for CLIP can further help improve the performance.

\begin{table}[!ht]
\centering\small
\caption{Overall accuracy on CIFAR10/100-LT with varying imbalance factors (IF). The best is in bold and the 2nd best is underlined. `-' indicates that the accuracy is not available in the original paper.}
\label{overall_cifar_accuracy}
\setlength\tabcolsep{4pt}
\begin{tabular}{l|ccc|ccc}
\toprule
Dataset & \multicolumn{3}{c|}{CIFAR10-LT} & \multicolumn{3}{c}{CIFAR100-LT}\\
\midrule
IF & 100 & 50 & 10 & 100 & 50 & 10\\
\midrule
BBN \cite{zhou2020bbn}
& 79.8 & 82.2 & 88.3 & 42.6 & 47.0 & 59.1 \\
LDAM \cite{NEURIPS2019_621461af}
& 77.0 & - & 88.2 & 42.0 & - & 58.7 \\
LiVT \cite{xu2023learning}
& 86.3 & -    & 91.3 & 58.2 & - & 69.2 \\
RIDE \cite{wang2021longtailed}
& - & -& -& 48.0 & 51.7 & 61.8\\
SHIKE \cite{jin2023longtailed}
& - & - & - & 56.3 & 59.8 & - \\
TADE \cite{zhang2022self}
& - & - & - & 49.8 & 53.9 & 63.6 \\
GLMC \cite{du2023global}
& 87.8 & 90.2 & 94.0 & 57.1 & 62.3 & 72.3 \\
MARC \cite{wang2023margin}
& 85.3 & - & - & 50.8 & - & - \\
\bottomrule
\rowcolor[gray]{.9}\multicolumn{7}{c}{CLIP (ViT-B/32)}\\
CE
& 89.8 & 90.0 & 91.6 & 67.5 & 68.1 & 70.4 \\
BalCE
& 91.3 & 91.6 & 92.4 & 67.6 & 68.8 & 70.8\\
Focal
& 89.8 & 90.0 & 91.6 & 66.9 & 68.6 & 70.4\\
LDAM
& 89.7 & 91.5 & 94.6 & 70.5 & 72.1 & 77.2\\
MMS
& \underline{93.3} & \underline{94.5} & 94.4 & 75.2 & 77.5 & 82.0\\
LFM + CE
& \textbf{93.8} & \textbf{95.2} & \underline{96.6} & \underline{77.3} & \underline{78.2} & \underline{82.6}\\
LFM + MMS
& 90.0 & 91.0 & \textbf{97.0} & \textbf{79.4} & \textbf{81.1} & \textbf{85.7}\\
\bottomrule
\end{tabular}
\end{table}

Then, we compare the fine-tuned CLIP models (ViT-B/32 is adopted) including our proposal with multiple existing state-of-art long-tailed learning methods in Table~\ref{overall_cifar_accuracy} under different imbalance factors. We can observe that by fine-tuning the pre-trained CLIP image encoder, the performance can be significantly improved in all scenarios. Moreover, the state-of-the-art imbalance loss MMS~\cite{shu2023clipood} is very helpful, while our proposal can further significantly improve the performance in most cases. This further demonstrates the proposal of alleviating the class imbalance problem using pre-trained vision-language model and the effectiveness of LFM. It should be noted that methods using the backbone of ResNet50 are performing worse in general, and thus ResNet50 is not adopted in the following experiments. 

\paragraph{ImageNet-LT and Places-LT}
We construct ImageNet-LT \cite{openlongtailrecognition} by forming a subset of the ImageNet 2014 dataset \cite{5206848}. The resulting imbalance ratio of ImageNet-LT is $256$. As shown in Table~\ref{imagenet_accuracy}, we can observe that compared to existing methods, our method begets better performance especially on few-shot accuracy (i.e., for tail classes) by both rebalancing and leveraging semantic similarities of classes. Observing that the LFM + MMS performance for minor classes falls behind LFM + CE, we hypothesize that MMS's sole focus on exercising semantic similarities and ignorance of class sample frequencies may overfit the many classes. A technique that only focuses on one may be problematic for tasks where semantic similarities happens to exist more frequently among the many classes.

\begin{table}[h]

\centering\small
\caption{Performance comparison on ImageNet-LT. The best is in bold and the 2nd best is underlined. `-' means not available in the original paper.}
\vspace{1em}
\label{imagenet_accuracy}
\begin{tabular}{l|cccc}
\toprule
Methods & Many & Med & Few & All\\
\midrule
CE \cite{cui2019class}
& 64.0 & 33.8 & 5.8 & 41.6 \\
LDAM \cite{NEURIPS2019_621461af}
& 60.4 & 46.9 & 30.7 & 49.8 \\
LiVT \cite{xu2023learning}
& \underline{76.4} & 59.7 & 42.7 & 63.8 \\
RIDE \cite{wang2021longtailed}
& 68.3 & 53.5 & 35.9 & 56.8 \\
SHIKE \cite{jin2023longtailed}
& - & - & - & 59.7 \\
TADE \cite{zhang2022self}
& 66.5 & 57.0 & 43.5 & 58.8 \\
GLMC \cite{du2023global}
& 70.1 & 55.9 & 45.5 & 57.2 \\
MARC \cite{wang2023margin}
& 60.4 & 50.3 & 36.6 & 52.3 \\
\bottomrule
\rowcolor[gray]{.9}\multicolumn{5}{c}{CLIP (ViT-B/16)}\\
Zero-shot
& 69.2 & 66.8 & \underline{65.8} & 67.6 \\
LFM + CE & 69.8 & \textbf{71.8} & \textbf{68.7} & \underline{70.6}\\
LFM + MMS & \textbf{79.7} & \underline{71.4} & 51.3 & \textbf{71.7}\\
\bottomrule
\end{tabular}
\end{table}

In addition, we conduct experiments on Places-LT \cite{openlongtailrecognition} using LFM with CE and MMS. Places-LT is a long-tailed subset of the original dataset Places2 \cite{zhou2017places}. It is a dataset for scene classification containing 365 classes, and it suffers from extreme imbalance ($\gamma = 996$). To account for its imbalance severity, we adjust local feature mixup hyperparameters to be highly in favor of the minority classes. We increase the value of $\tau$, so that the probability distribution constructed by local sampling is more balanced. Additionally, we increase the value of $\alpha$, so that the label is shifted to the tail classes, more heavily as shown in the supplementary. 

\begin{table}[h]

\centering\small
\caption{Performance comparison on Places-LT. The best is in bold and the 2nd best is underlined. `-' indicates that the accuracy is not available in the original paper.}
\vspace{1em}
\label{places_accuracy}
\begin{tabular}{l|cccc}
\toprule
Methods & Many & Med & Few & All\\
\midrule
CE \cite{cui2019class}
& \underline{45.7} & 27.3 & 8.2 & 30.2 \\
Focal \cite{lin2017focal}
& 41.1 & 34.8 & 22.4 & 34.6 \\
LiVT \cite{xu2023learning}
& \textbf{50.7} & 42.4 & 27.9 & \underline{42.6} \\
SHIKE \cite{jin2023longtailed}
& 43.6 & 39.2 & 44.8 & 41.9 \\
TADE \cite{zhang2022self}
& 43.1 & 42.4 & 33.2 & 40.9 \\
MARC \cite{wang2023margin}
& 39.9 & 39.8 & 32.6 & 38.4 \\
\bottomrule
\rowcolor[gray]{.9}\multicolumn{5}{c}{CLIP (ViT-B/16)}\\
Zero-shot
& 36.8 & 35.8 & 45.1 & 38.1 \\

LFM + CE
& 41.3 & \underline{43.5} & \underline{46.2} & 42.3\\
LFM + MMS
& 45.2 & \textbf{48.5} & \textbf{46.6} & \textbf{46.9}\\
\bottomrule
\end{tabular}
\end{table}

Table~\ref{places_accuracy} summarizes the results. The benefit from the pre-trained model by CLIP can be observed from the zero-shot performance on tail classes, which further demonstrates the advantage of the text supervision from CLIP. However, fine-tuning using our proposal is necessary to improve the performance. It should also be noted that due to the severe imbalance factor of this data, our proposal with CE is expected to be less effective compared to that with MMS~\cite{shu2023clipood}. LFM with MMS shows significantly better performance compared to state-of-the-arts, especially on medium-shot and few-shot classes, and demonstrates strong performance on many-shot classes as well. This further demonstrates the effectiveness of our proposed method on the long-tailed problem.

\paragraph{Effect of Mixup Techniques}
To demonstrate the proposed LFM, we also compare it with the standard Mixup~\cite{zhang2018mixup} and Remix~\cite{chou2020remix} on CIFAR100-LT. Specifically, we fine-tune CLIP with the same hyperparameters and decoupled stages, using different mixup techniques. Each model is trained using cross entropy loss with the ViT-B/16 backbone. Remix is a mixup method that addresses the class-imbalance issue, and it makes a trade-off between many-shot and few-shot performances. For example, compared to the standard mixup, Remix can help improve the performance on few-shot classes but sacrifice the performance on many-shot classes. However, Remix ignores the semantic relationship between each class pair that CLIP can be used for. Our proposal shows significantly better performance in Table~\ref{mixup_ablation}.

\begin{table}[ht]
\centering
\caption{Comparison of different mixup techniques on CIFAR100-LT with imbalance factor of 100.}
\label{mixup_ablation}
\begin{tabular}{l|cccc}
\toprule
Methods & Many & Med & Few & All\\
\midrule
Mixup \cite{zhang2018mixup}
& 80.4 & 71.5 & 55.1 & 69.5\\
Remix \cite{chou2020remix}
& 79.6 & 71.5 & 55.7 & 69.4\\
LFM 
& \textbf{83.4} & \textbf{83.3} & \textbf{72.8} & \textbf{80.1}\\
\bottomrule
\end{tabular}
\end{table}

\begin{figure*}[h]
    \begin{subfigure}{0.15\textwidth}
        \includegraphics[width=\textwidth]{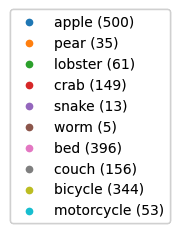}
    \end{subfigure}
    \begin{subfigure}{0.27\textwidth}
        \includegraphics[width=\textwidth]{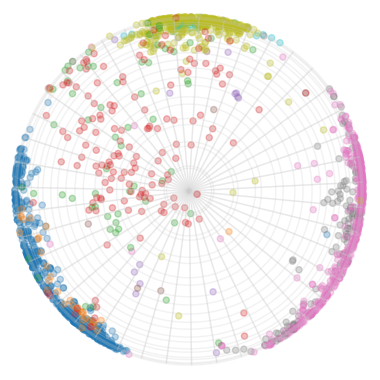}
        \caption{Zero Shot}\label{top_feat}
    \end{subfigure}
    \hfill
    \begin{subfigure}{0.27\textwidth}
        \includegraphics[width=\textwidth]{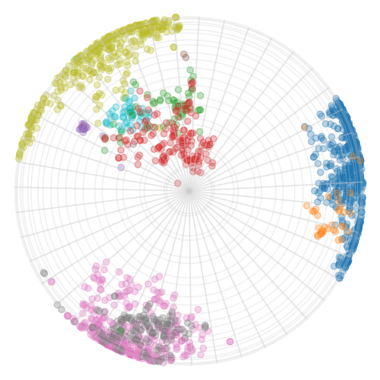}
        \caption{CE}\label{top_feat_ce}
     \end{subfigure}
     \hfill
    \begin{subfigure}{0.27\textwidth}
        \includegraphics[width=\textwidth]{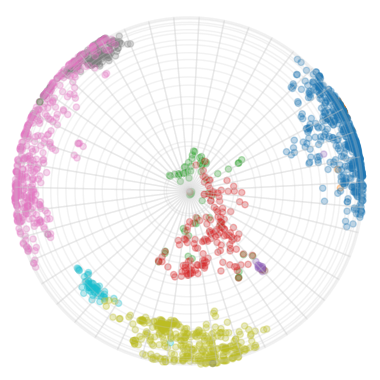}
        \caption{CE + LFM}\label{top_feat_lfm_ce}
     \end{subfigure}
    \caption{An illustration on image distribution of sampled classes using feature vectors extracted from different image encoders.}\label{TOP_FEATURES}
\end{figure*}

\paragraph{Visualization}
To demonstrate our assumption on the local semantic relationship, we illustrate the geometric effect of fine-tuning CLIP with our proposal in Fig.~\ref{TOP_FEATURES}. We demonstrate the effect by revealing the contrastive locality of image feature vector outputs, where the input is comprised of a set of randomly sampled images from 10 chosen classes $\{$apple, pear, $\dots$, motorcycle$\}$ in CIFAR100. Our illustration contains the following 5 pairs of semantically related categories from CIFAR100: (apple, pair), (lobster, crab), (snake, worm), (bed, couch), and (bicycle, motorcycle). The legend contains the name of the class and the sample count in parenthesis. We choose these pairs to show that their semantic relations are aligned with their visual relations in terms of contrastive locality, as perceived by the image encoding layers.

With the ViT-B/32 vision encoder and fully connected layer, we obtain a 512-dimensional feature vector for each image. To reduce the high-dimensional feature vectors to three dimensions for human readability, we convert them using t-SNE trained for 1000 iterations and seed set to 1. At zero shot, we can observe that semantically related classes are located nearby (e.g., `apple' vs. `pear' and `bed' vs. `couch'), although some are poorly clustered (e.g., `lobster' vs. `crab'). This confirms our assumption that pre-trained vision-language model can align semantically related classes together. However, the separation between non-related classes are not clear in the pre-trained model. By fine-tuning the image encoder with cross entropy loss, the separation between non-related classes becomes clear, thanks to the help of head-class training data. However, we can observe that tail-class instances are largely overlapping with semantically related head-class instances (e.g., `lobster' vs. `crab'). Fortunately, by incorporating our proposal of LFM, tail-class instances can be pushed a bit away from their semantically related head-class instances without sacrificing the clear boundaries between non-related classes, which further demonstrates our proposal. 

\section{Conclusion}
Considering CLIP's ability to generalize to unseen categories, we leverage a fixed text encoder to enhance the performance of image categorization over long-tailed training distributions.

We enable the accuracy boost with the construction of a novel mixup technique that takes advantage of the semantic relationships between classes by probabilistic sampling based on their locality in the text encoder's feature space and slightly shift the label towards tail classes. Our extensive experiments on several benchmark long-tailed training data demonstrate the effectiveness of our proposal in alleviating the class imbalance issue with an efficient strategy that incorporates a fixed text encoder. Local feature mixup can be easily applied to not only vision-language backbones but also non multi-modal methods (i.e. vision-only architectures), which will be studied in our future work. However, both LFM and vision-language image classification are limited by the domain knowledge of the text encoder. Without further training, pre-trained CLIP performs poorly on domain-specific tasks as suggested by \cite{tian2022vl, dong2023lpt, jia2022visual} due to its generic knowledge. 
Our method relies on the ability of the text encoder to capture pairwise semantic similarities among the class names present in the dataset which proves to be performant for the common domain such as CIFAR and ImageNet but not for biological names present in iNaturalist \cite{van2018inaturalist}, which will be our future work.  

\section{Acknowledgement}
Yao and Hu's research is supported in part by NSF (IIS-2104270) and Advata Gift Funding. Zhong's research is supported in part by the Carwein-Andrews Graduate Fellowship and Advata Gift Funding. All opinions, findings, conclusions and recommendations in this paper are those of the author and do not necessarily reflect the views of the funding agencies.

\bibliography{bib}

\appendix
\section{Appendix}
\subsection{Effect of Local Sampling}
\label{local_sampling_distribution}

\begin{figure*}[hb]
    \begin{subfigure}{0.32\textwidth}
        \centering
        \includegraphics[width=\textwidth]{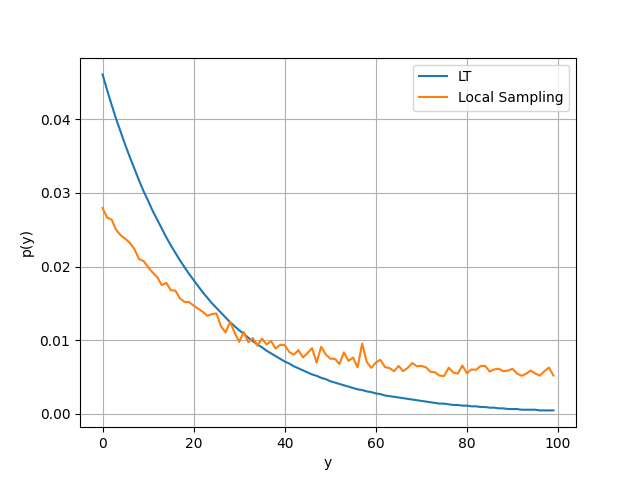}
        \caption{$\gamma=100$, $\gamma' = 5.44$}
        \label{CIFAR100_LS_100}
    \end{subfigure}
    \hfill
    \begin{subfigure}{0.32\textwidth}
        \centering
        \includegraphics[width=\textwidth]{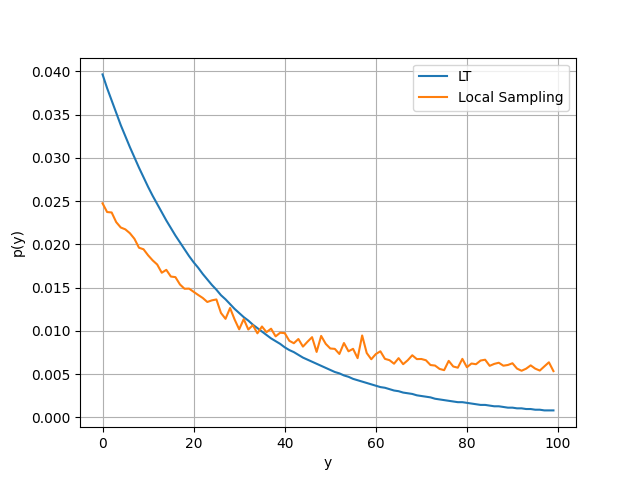}
        \caption{$\gamma=50$, $\gamma' = 4.64$}
        \label{CIFAR100_LS_50}
     \end{subfigure}
     \hfill
    \begin{subfigure}{0.32\textwidth}
        \centering
        \includegraphics[width=\textwidth]{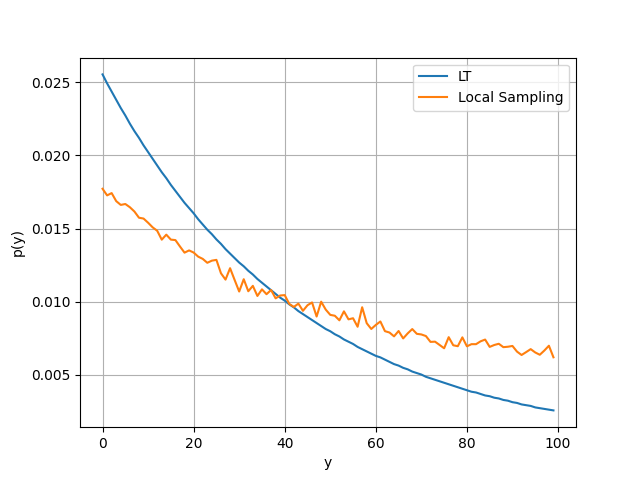}
        \caption{$\gamma=10$, $\gamma' = 2.86$}
        \label{CIFAR100_LS_10}
     \end{subfigure}
    \caption{Local sampling effect on CIFAR100-LT~\cite{krizhevsky2009learning} $p(y)$ distribution}
    \label{CIFAR100_LS}
\end{figure*}
\begin{figure*}[hb]
    \begin{subfigure}{0.32\textwidth}
        \centering
        \includegraphics[width=\textwidth]{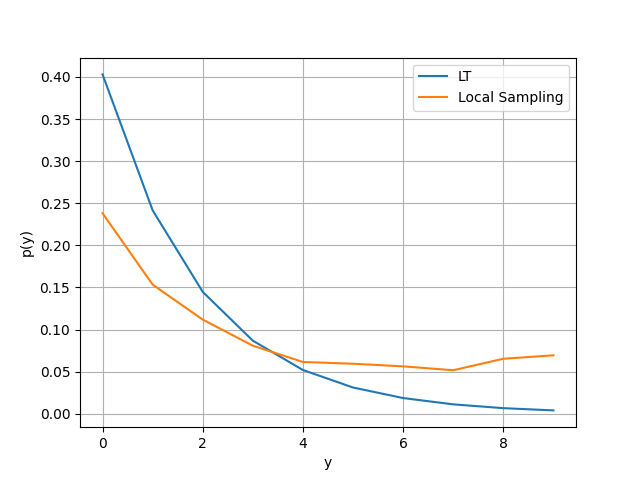}
        \caption{$\gamma=100$, $\gamma' = 4.60$}
        \label{CIFAR10_LS_100}
    \end{subfigure}
    \hfill
    \begin{subfigure}{0.32\textwidth}
        \centering
        \includegraphics[width=\textwidth]{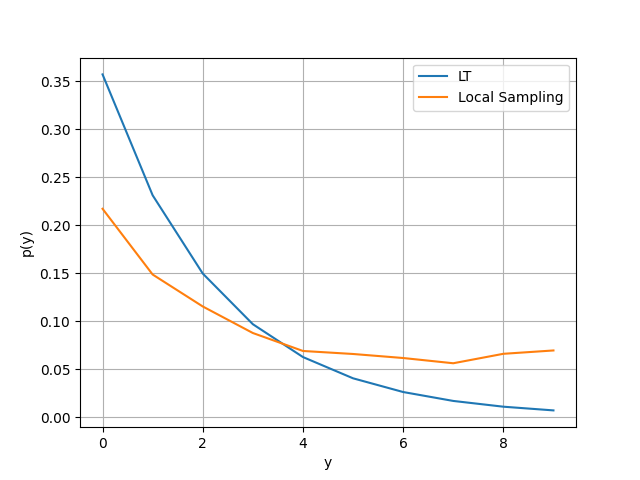}
        \caption{$\gamma=50$, $\gamma' = 3.86$}
        \label{CIFAR10_LS_50}
     \end{subfigure}
     \hfill
    \begin{subfigure}{0.32\textwidth}
        \centering
        \includegraphics[width=\textwidth]{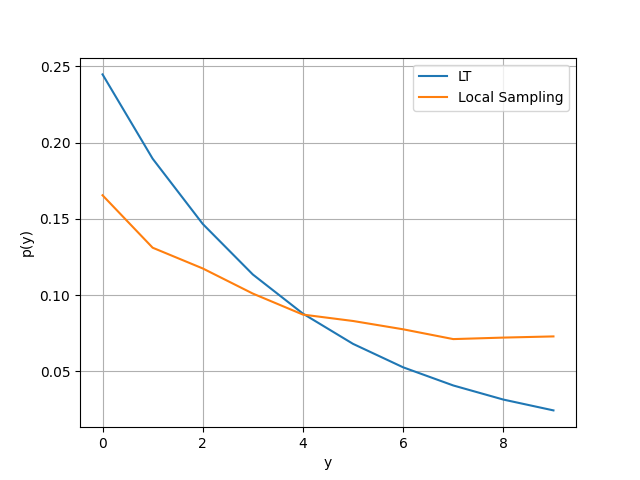}
        \caption{$\gamma=10$, $\gamma' = 2.32$}
        \label{CIFAR10_LS_10}
     \end{subfigure}
    \caption{Local sampling effect on CIFAR10-LT~\cite{krizhevsky2009learning} $p(y)$ distribution}
    \label{CIFAR10_LS}
\end{figure*}
\begin{figure*}[hb]
    \centering
    \begin{subfigure}{0.4\textwidth}
        \centering
        \includegraphics[width=\textwidth]{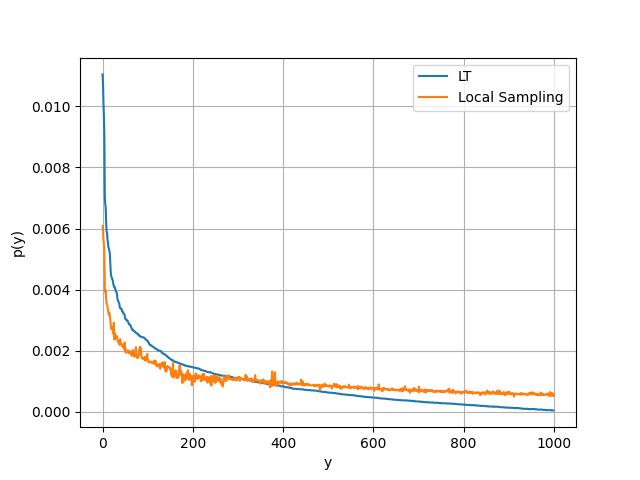}
        \caption{ImageNet-LT ($\gamma=256$, $\gamma' = 12.22$)}
        \label{ImageNet_LS}
    \end{subfigure}
    \begin{subfigure}{0.4\textwidth}
        \centering
        \includegraphics[width=\textwidth]{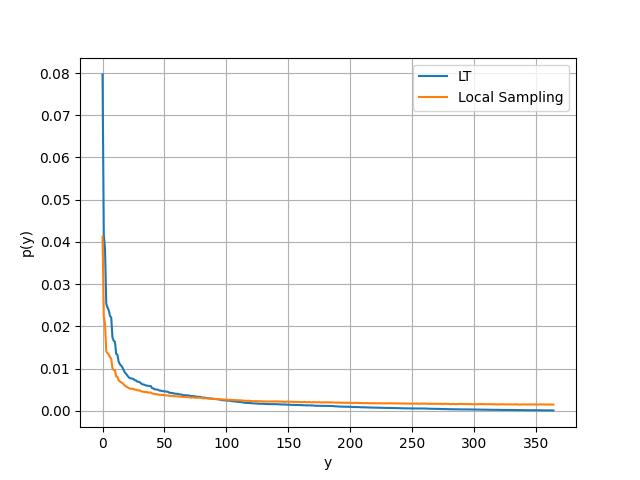}
        \caption{Places-LT ($\gamma=996$, $\gamma' = 28.59$)}
        \label{Places_LS}
     \end{subfigure}
     \caption{Local sampling effect on ImageNet-LT~\cite{openlongtailrecognition} and Places-LT~\cite{zhou2017places} $p(y)$ distribution}\label{ImageNet_Places_LT}
\end{figure*}

As discussed in the main paper, at each training step, local sampling feeds the model an image pair that holds semantically-related images, where the semantic relation is determined by the text encoder. In constructing the pair, the label of the first image is determined by 
\begin{align}
\label{p_i}
    p(y = y_i) = \frac{n_i}{\sum^C_{k}n_k}
\end{align}
which is to uniformly sample an image without replacement from the dataset. However, the label of the second image is determined by
\begin{equation}
\label{p_j_i}
p_{ls}(y = y_j | y_i) = 
    \begin{cases}
        \frac{ exp( f_{T_i} \cdot f_{T_j} / \tau ) }{ \sum^{C}_{k=1} exp( f_{T_i} \cdot f_{T_k} / \tau ) } & \text{ } i \neq j\\
        0 & \text{o.w.}\
    \end{cases}
\end{equation}
which ignores the sample count for any class label. Due to the negligence of the second label's sample count, the amount of times that the model sees minority classes can be increased effectively balancing the data distribution by resampling. To observe the amount of resampling, we show the sample count before and after local sampling as follows. Allow $Y$ to be the random variable in the event that local sampling yields an instance of class $y \in \{y_i, y_j\}$, and allow $y_i$ to be the event that $y_i = y$ and $y_j$ to be the event that $y_j = y$. The probability that the model observes an image with class label $y$ can be calculated as
\begin{align*}
    p(Y=y) &= p(y_i) + (1 - p(y_i)) p(y_j)\\
    &= p(y_i) + (1 - p(y_i))\sum_{k, k\neq i}^C p(y_j | y_k)p(y_k) \text{ .}
\end{align*}
Using Eqns. \ref{p_i} and \ref{p_j_i}, $p(Y)$ can be evaluated for all $y$, and we illustrate the resulting $p(y)$ for every dataset in Figs.~\ref{CIFAR100_LS}-\ref{ImageNet_Places_LT}. Additionally, we indicate the new imbalance factor as $\gamma'$. We can observe that the imbalance severity and the magnitude of long-tailed distribution can be well reduced, which demonstrates the effectiveness of our local sampling method. 

\subsection{Comparison between Textual Similarity and Visual Categorization}

\begin{figure*}[h]
    \centering
    \begin{subfigure}[b]{0.45\textwidth}
        \includegraphics[width=\linewidth]{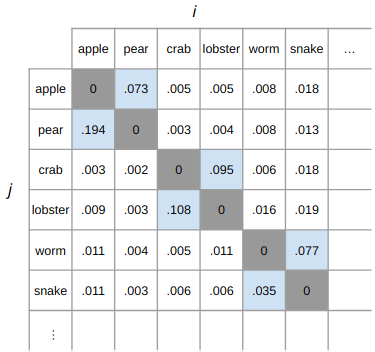}
        \caption{$p_{ls}(y=y_j|y_i)$}
        \label{prob_dist}
    \end{subfigure}
    \hspace{0.05\textwidth}
    \begin{subfigure}[b]{0.45\textwidth}
        \includegraphics[width=\linewidth]{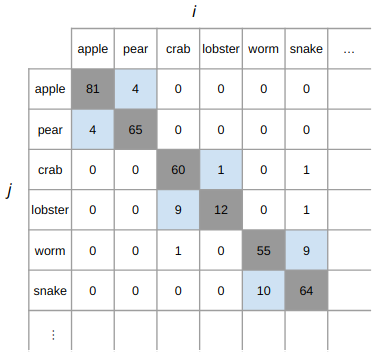}
        \caption{Confusion Matrix}
        \label{confusion_matrix}
    \end{subfigure}
    \caption{The tables above demonstrate the correlation between text feature similarities (captured by $p_{ls}$) and the model performance with zero-shot classification. The diagram on the left shows our constructed probability distribution $p_{ls}$ for CIFAR100~\cite{krizhevsky2009learning}, and the diagram on the right is a confusion matrix of CLIP's performance on CIFAR100 without training. The columns represent class $y_i$, and the rows represent class $y_j$. For demonstration purposes, we present three pairs of related classes: (apple, pear), (crab, lobster), and (snake, worm). Blue cells hold values for related class pairs while gray cells can be ignored since they hold the values for same class pairs. It can be observed that the blue cells hold values that are generally higher than any of the other white cells in their respective rows.}
    \label{sample_prob_dist}
\end{figure*}

\begin{figure}[b!]
    \begin{subfigure}{.5\textwidth}
        \centering
        \includegraphics[width=0.5\textwidth]{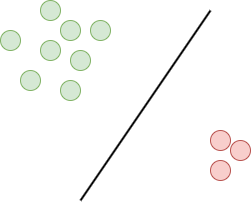}
        \caption{$\alpha = 0$}
        \label{lfm_effect_a}
    \end{subfigure}
    \hfill
    \begin{subfigure}{.5\textwidth}
        \centering
        \includegraphics[width=0.5\textwidth]{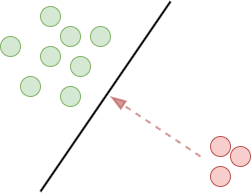}
        \caption{$\alpha > 0$}
        \label{lfm_effect_b}
     \end{subfigure}
    \caption{An illustration of the theorized effect that label shift has on the model's decision boundary. Red circles indicate feature vectors of tail classes and green circles are that of nearby head classes. When $\alpha > 0$, the decision boundary shifts towards the head classes anticipating for higher intra-class variance for tail classes.}
    \label{lfm_effect}
     \vspace{-4mm}
\end{figure}

To further confirm our assumption that semantically related classes are visually related, we make a comparison between class label textual similarities and CLIP's zero-shot performance. Fig.~\ref{sample_prob_dist} shows a comparison between our semantic probability distribution $p_{ls}$ and a confusion matrix of CLIP's zero-shot classification performance using CIFAR100's validation set. It can be observed that $p_{ls}$ is correlated with the performance of zero-shot classification. By observing the blue cells in the confusion matrix, we see that the model more frequently struggles to find a decision boundary between related classes. When we sample with $p_{ls}$, we expect that we are sharing information with related classes more frequently and thus establish a decision boundary more optimally positioned for inference on the balanced validation data.

\subsection{Algorithms and Training Configurations}

In this section, we summarize the algorithms for LocalSample, Mix, and the entire training process, where the effect on the proposed mixup technique on the decision boundary between nearby head and tail classes is illustrated in Fig.~\ref{lfm_effect}. Upon acceptance of this paper, we will also publicly release the code.

\begin{algorithm}[h]
\caption{LocalSample ($\tau$, $f_T$, $D = \{(x, y)\}$)}\label{alg:localsample}
\begin{algorithmic}[1]
\State $p_{y_i} \gets [0,1]^{C}$ vector representing the probability distribution from Eqn. \ref{p_i}
\State $p_{y_j|y_i} \gets [0,1]^{C\times C}$ matrix representing the probability distribution from Eqn. \ref{p_j_i} with given $\tau$ and $f_T$

\While{model is not converged}
    \State $y_i \sim p_{y_i}$
    \State $y_j \sim p_{y_j|y_i}$
    \State $x_i \sim \{ x \mid (x, y) \in D \text{ and } y = y_i \}$
    \State $x_j \sim \{ x \mid (x, y) \in D \text{ and } y = y_j \}$
    \State \textbf{yield} $(x_i, y_i)$, $(x_j, y_j)$ 
\EndWhile
\end{algorithmic}
\end{algorithm}

\begin{algorithm}[h]
\caption{Mix ($\alpha$, $(x_i, y_i)$, $(x_j, y_j)$)}\label{alg:mix}
\begin{algorithmic}[1]
    \State Convert $y_i$ and $y_j$ to one-hot vectors of size $C$
    \State $\lambda_x \sim Beta(0.5, 0.5)$
    \State $\lambda_y \gets $ label shift assignment by Eqn. 2 in the main paper 
    \State $x^{LFM} \gets \lambda_x x_i + (1 - \lambda_x) x_j$
    \State $y^{LFM} \gets \lambda_y y_i + (1 - \lambda_y) y_j$
    \State \Return $x^{LFM}$, $y^{LFM}$
\end{algorithmic}
\end{algorithm}

\begin{algorithm}[h]
\caption{Train ($\mathscr{F}_T$,  $\mathscr{F}_I$, $W_I$, $\alpha$, $\tau$, $D$, $T$)}
\begin{algorithmic}[1]
   \State Initialize $\Theta_T$ and $\Theta_I$ (weights of $\mathscr{F}_T$ and $\mathscr{F}_I$, respectively) with pre-trained weights
   \State Freeze $\Theta_T$
   \State $f_T \gets \Pi_{||.||_2=1}\mathscr{F}_T(T)$
   \For{epoch in $1,\dots, N_0$} \Comment{Stage 1}
       \For{$(x_i, y_i), (x_j, y_j)$ in LocalSample ($\tau$, $f_T$, $D$)}
           \State $x^{LFM}, y^{LFM} \gets Mix (\alpha, (x_i, y_i), (x_j, y_j))$
           \State $f_I \gets \Pi_{||.||_2=1}\mathscr{F}_I(x^{LFM})$
           \State $\ell \gets \mathcal{L}(f_T \cdot f_I, y^{LFM})$
           \State Update $\Theta_{I}$
       \EndFor
   \EndFor
   \State Freeze $\Theta_I$ \Comment{Stage 2}
   \State Initialize $W_I$ as $d \times d$ identity matrix, $I_d$, where $d$ is the feature dimension of $\mathscr{F}_I$
   \For{epoch in $1,\dots, N_1$}
       \For{$(x_i, y_i), (x_j, y_j)$ in LocalSample ($\tau$, $f_T$, $D$)}
           \State $x^{LFM}, y^{LFM} \gets Mix (\alpha, (x_i, y_i), (x_j, y_j))$
           \State $f_I \gets \Pi_{||.||_2=1}(W_I^T \mathscr{F}_I(x^{LFM}))$
           \State $\ell \gets \mathcal{L}(f_T \cdot f_I, y^{LFM})$
           \State Update $W_I$
       \EndFor
   \EndFor
\end{algorithmic}
\end{algorithm}

During the training, we use the hyperparameters and other training properties listed in Table~\ref{hyperparameters}. Most experiments have the same setup, but some minor adjustments are made largely due to differences in class label distributions. Under the circumstances of heavy class imbalance, we can simply raise the values of $\alpha$ and $\tau$, which we do for Places-LT~\cite{zhou2017places}. Detailed information for each dataset is provided in Table~\ref{datasets}. The original dataset imbalance is summarized by the imbalance factor $\gamma$.

\subsection{Additional Ablation Studies}

Besides the ablation study conducted in the main paper, we also conducted the following ablation studies.

\begin{table*}[t!]
\caption{Hyperparameters and configurations.}
\vspace{1em}
\label{hyperparameters}
\centering
\resizebox{\linewidth}{!}{
\begin{tabular}{l|cc|cc|cc|cc}
\toprule
Dataset & 
\multicolumn{2}{c|}{CIFAR10-LT~\cite{krizhevsky2009learning}}& 
\multicolumn{2}{c|}{CIFAR100-LT~\cite{krizhevsky2009learning}}& 
\multicolumn{2}{c|}{ImageNet-LT~\cite{openlongtailrecognition}}&
\multicolumn{2}{c}{Places-LT~\cite{zhou2017places}}\\

\midrule

Stage & 
1 & 2 & 1 & 2 & 1 & 2 & 1 & 2\\
\midrule
Epochs & 
10 & 10 & 50 & 10 & 30 & 10 & 30 & 10\\
Learning Rate & 
$1\times 10^{-9}$ & $5\times 10^{-1}$ & $1\times 10^{-6}$ & $1\times 10^{-2}$ & $5\times10^{-6}$ & $1\times 10^{-2}$ & $1\times10^{-7}$ & $5\times10^{-4}$\\
LR Scheduler & 
\multicolumn{2}{c|}{Cosine Annealing} & \multicolumn{2}{c|}{Cosine Annealing} & \multicolumn{2}{c|}{Cosine Annealing} & \multicolumn{2}{c}{Cosine Annealing}\\
Min LR & 
$1\times 10^{-12}$ & $5\times 10^{-4}$ & $1\times 10^{-9}$ & $1\times 10^{-5}$ & $5\times10^{-9}$& $1\times 10^{-5}$ & $1\times10^{-10}$& $5\times10^{-7}$\\
Optimizer & 
\multicolumn{2}{c|}{Adam} & \multicolumn{2}{c|}{Adam}& \multicolumn{2}{c|}{Adam}& \multicolumn{2}{c}{Adam}\\
Batch Size & 
\multicolumn{2}{c|}{32} & \multicolumn{2}{c|}{32}& \multicolumn{2}{c|}{96} & \multicolumn{2}{c}{96}\\
$\alpha$ for LFM & 
\multicolumn{2}{c|}{1.00} & \multicolumn{2}{c|}{1.00}& \multicolumn{2}{c|}{1.00} & 1.25 & 1.50\\
$\tau$ for LFM & 
\multicolumn{2}{c|}{0.05} & \multicolumn{2}{c|}{0.05}& \multicolumn{2}{c|}{0.05}& \multicolumn{2}{c}{1.00}\\
Seed & 
\multicolumn{2}{c|}{0} & \multicolumn{2}{c|}{0}& \multicolumn{2}{c|}{0}& \multicolumn{2}{c}{0}\\
\bottomrule
\end{tabular}
}
\end{table*}

\begin{table*}[t!]
\caption{Detailed information of mentioned datasets}
\vspace{1em}
\label{datasets}
\centering
\resizebox{\linewidth}{!}{
\begin{tabular}{l|ccc|ccc|c|c}
\toprule
Dataset & 
\multicolumn{3}{c|}{CIFAR10-LT~\cite{krizhevsky2009learning}}& 
\multicolumn{3}{c|}{CIFAR100-LT~\cite{krizhevsky2009learning}}& 
\multicolumn{1}{c|}{ImageNet-LT~\cite{openlongtailrecognition}}&
\multicolumn{1}{c}{Places-LT~\cite{zhou2017places}}\\
\midrule
Number of classes &
\multicolumn{3}{c|}{10} & \multicolumn{3}{c|}{100} & 1000 & 365\\
Total Training Images &
20,431 & 13,996 & 12,406 & 19,573 & 12,608 & 10,847 & 115,846 & 62,500\\
Max Images &
5,000 & 5,000 & 5,000 & 500 & 500 & 500 & 1,280 & 4,980\\
Min Images &
500 & 100 & 50 & 50 & 10 & 5 & 5 & 5\\
Original Imbalance Factor $\gamma$ &
10 & 50 & 100 & 10 & 50 & 100 & 256 & 996\\
Effective Imbalance Factor $\gamma '$ &
2.32 & 3.86 & 4.60 & 2.86 & 4.64 & 5.44 & 12.22 & 28.59\\

\bottomrule
\end{tabular}
}
\end{table*}

\begin{figure}[t!]
    \centering
    \begin{minipage}{0.49\textwidth}
    \centering
    \includegraphics[width=\textwidth]{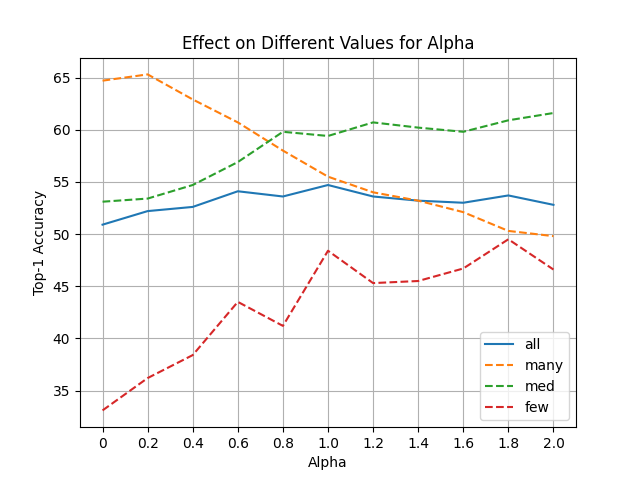}
    \caption{Effect of different $\alpha$.}
    \label{alpha_ablation}
    \end{minipage}
    \begin{minipage}{0.49\textwidth}
    \centering
    \includegraphics[width=\textwidth]{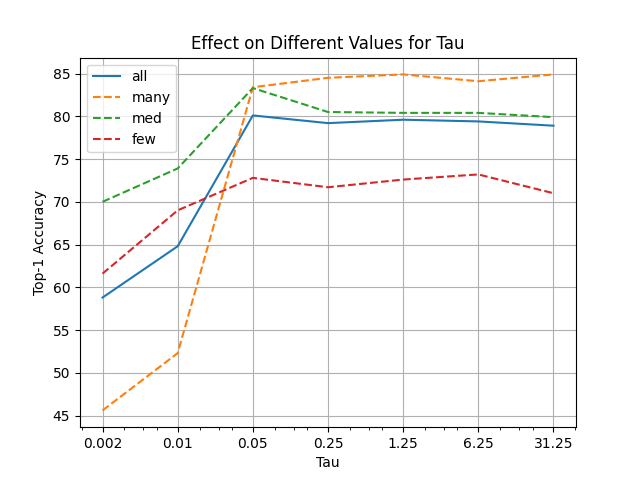}
    \caption{Effect of different $\tau$.}
    \label{tau_ablation}
    \end{minipage}
\end{figure}

\subsubsection{Effect of $\alpha$}
We study the effect of the intensity in which we shift the training label assigned to each mixup, for which we can control with $\alpha$. The $\alpha$ value directly affects the positioning of the model's decision boundaries between class pairs, and we can expect lower values to extend the boundary of many-shot classes and higher values to extend the boundary of few-shot classes. In this study, we change $\alpha$ among the range $[0, 2]$ on CIFAR100-LT with an imbalance factor of 100 using CLIP's ResNet50 backbone with the same configuration settings. From Fig.~\ref{alpha_ablation}, we can easily observe that an increasing of $\alpha$ can slowly degenerate the performance of many-shot classes while improve the performance of the other, especially that of the few-shot classes as expected. The result also reveals that setting $\alpha$ to 1 works best for all accuracies.

\subsubsection{Effect of $\tau$}
To study the effect of different temperature settings for $p_{ls}$, we run multiple experiments with $\tau=\{.002, .01, .05, .25, 1.25, 31.25\}$. At lower values, we increase the probability that nearby class samples $(i,j)$ are paired together. At higher values, the probability of two nearby class samples becoming paired is mitigated, and the class sampling becomes more balanced. We run our experiments on CIFAR100-LT~\cite{krizhevsky2009learning} with an imbalanced factor of 100 using CLIP's ViT-B/16 backbone, which is of the same configuration settings. 
Fig.~\ref{tau_ablation} reveals that when we increase $\tau$ from a small value, all classes can benefit from LFM by mixing semantically related samples, while after $\tau=0.05$ it plateaued. Therefore, $\tau=0.05$ is adopted in the rest of our experiments.

\end{document}